\documentclass[letterpaper, 10 pt, conference]{ieeeconf}

\usepackage[T1]{fontenc}
\usepackage{latexml}

\IEEEoverridecommandlockouts
\usepackage{amsmath,amssymb}
\usepackage{graphicx}
\makeatletter
\let\NEXUSoriginalmakecaption\@makecaption
\makeatother
\usepackage{caption}
\makeatletter
\AtBeginDocument{\let\@makecaption\NEXUSoriginalmakecaption}
\makeatother
\usepackage{booktabs}
\usepackage{array}
\usepackage[table]{xcolor}
\usepackage{cite}
\usepackage{url}
\usepackage{needspace}
\makeatletter
\let\NAT@parse\undefined
\makeatother
\definecolor{paperlinkblue}{RGB}{35,70,120}
\usepackage[colorlinks=true,linkcolor=paperlinkblue,citecolor=paperlinkblue,urlcolor=paperlinkblue]{hyperref}

\newcommand{\projectwebsite}{Project website: \url{https://nexus-humanoid.github.io/}.}

\title{\LARGE \bf
NEXUS: Perceptive Whole-Body Control\\
for Terrain-Adaptive Teleoperation
}

\author{\authorblockN{Xiangyu Miao$^{1,2}$, Junsong Wu$^{1,2}$, Jiyuan Shi$^{1}$, Weiji Xie$^{1,2}$, Jinrui Han$^{3}$,\\
Xingyi Wang$^{1,4}$, Weinan Zhang$^{2,\dagger}$, Chenjia Bai$^{1,5}$, Xuelong Li$^{1,\dagger}$}
\authorblockA{
{\fontsize{10}{12}\selectfont $^{1}$The Institute of Artificial Intelligence, China Telecom (TeleAI)\quad $^{2}$Shanghai Jiao Tong University}\\
{\fontsize{10}{12}\selectfont $^{3}$The University of Hong Kong\quad $^{4}$Zhejiang University\quad $^{5}\gamma$-Robotics}%
\iflatexml
\\\textsuperscript{\dag}Corresponding authors: Weinan Zhang (\href{mailto:wnzhang@sjtu.edu.cn}{\texttt{wnzhang@sjtu.edu.cn}}) and Xuelong Li (\href{mailto:xuelong_li@ieee.org}{\texttt{xuelong\_li@ieee.org}}).
\fi%
}}

\begin{document}

\twocolumn[{%
\renewcommand\twocolumn[1][]{#1}%
\maketitle
\begingroup
\begin{center}
  \centering
  \captionsetup{type=figure,font=footnotesize,labelsep=period}
  \includegraphics[width=\textwidth]{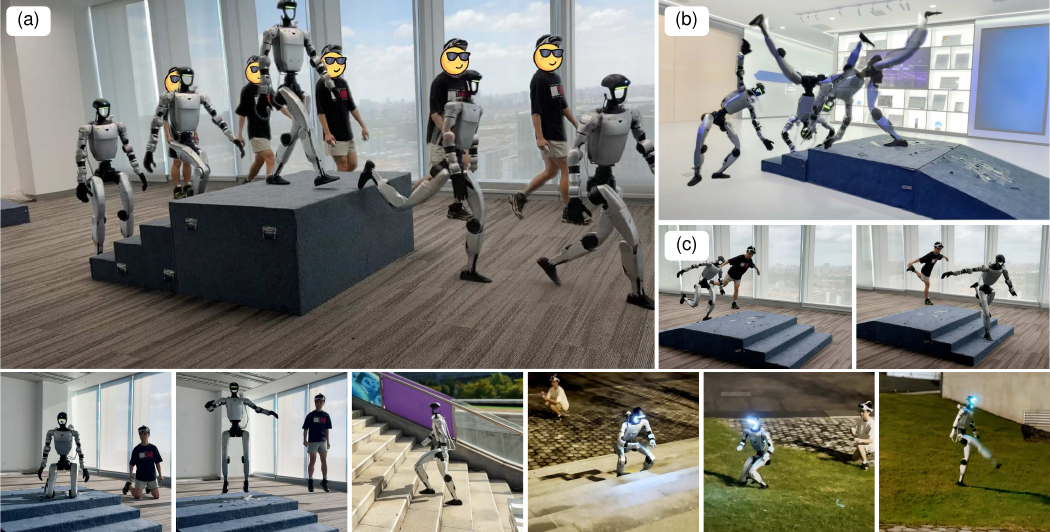}
  \caption{\textbf{Whole-body teleoperation under human--robot terrain mismatch with NEXUS.} The human operator performs motions on flat ground while the robot adapts the commanded behaviors to its local terrain. (a) Stair ascent followed by stepping down from a 60-cm-high platform. (b) Cartwheel across a platform edge. (c) From left to right, top row: single-leg balancing on a ramp and a step, using opposite support legs; bottom row: double-knee kneeling and vertical jumping on indoor platforms, stair climbing and squatting, and single-knee kneeling and kicking on grassy slopes.}
  \label{fig:teaser}
\end{center}
\endgroup

}]
\iflatexml
\else
\begingroup
\renewcommand{\thefootnote}{\fnsymbol{footnote}}
\footnotetext[2]{Corresponding authors: Weinan Zhang (\href{mailto:wnzhang@sjtu.edu.cn}{\texttt{wnzhang@sjtu.edu.cn}}) and Xuelong Li (\href{mailto:xuelong_li@ieee.org}{\texttt{xuelong\_li@ieee.org}}).}
\endgroup
\fi
\thispagestyle{empty}
\pagestyle{empty}

\begin{abstract}
Whole-body teleoperation requires a humanoid robot to reproduce a human operator's behavior even when their terrains differ.
This demands that the robot perceive local terrain and adapt its posture and contacts accordingly, rather than copy the operator's motion frame by frame.
However, paired motion data linking the same behaviors across flat ground and different terrains remain scarce, limiting supervision for learning terrain-adaptive control.
To enable whole-body teleoperation across mismatched terrains, we introduce NEXUS, a perceptive whole-body control framework that combines human motion commands with onboard sensory feedback.
We first develop a scalable terrain-aware adaptation algorithm that efficiently generates high-quality motion pairs across motions and terrains without per-motion or per-terrain tuning.
Using a paired motion corpus totaling nearly 1,000 hours, we train a perceptive whole-body controller through teacher--student learning to reproduce commanded behaviors across terrains.
Experiments demonstrate efficient, scalable generation of high-quality motion data and show that NEXUS combines broad behavioral coverage with terrain adaptability and tracking fidelity, outperforming existing whole-body controllers on the evaluated benchmarks.
Zero-shot real-world deployment enables real-time whole-body teleoperation on diverse unseen terrains, further validating the generalization of our method.
\projectwebsite
\end{abstract}

\section{Introduction}
\label{sec:introduction}

Whole-body teleoperation enables a humanoid robot to act as a human operator's physical proxy, reproducing diverse and coordinated body movements at a remote site~\cite{he2025omnih2o,ze2025twist,zeTWIST2ScalablePortable2025a,liu2026heft,sun2026mosaic,li2025clone,wu2026teleopit,fu2025humanplus,ben2025homie,li2026omniclone}.
Extending this capability beyond controlled flat-ground settings would allow robots to operate outdoors and perform tasks that involve traversing varied terrain.
In such settings, the operator may remain on flat ground while the robot encounters stairs, slopes, or other uneven surfaces.
Preserving the intended behavior then requires the robot to adapt its posture and contacts to local support conditions.
Such teleoperation demands control that combines a broad repertoire of behaviors with environmental adaptability.

Reinforcement learning has enabled legged robots to acquire locomotion and specialized motor skills~\cite{hwangbo2019agile,miao2025pala,han2026husky,kumar2021rma,rudin2022minutes,huang2025host}.
Perceptive locomotion handles challenging terrain through navigation goals or velocity commands~\cite{zhu2026hiking,zhang2026rpl,wang2026more,wang2025beamdojo,he2025ame,longLearningHumanoidLocomotion2024a}, but does not specify the coordinated body movements of a desired behavior.
Motion tracking provides this richer behavioral interface, enabling dynamic whole-body skills~\cite{pengDeepMimicExampleGuidedDeep2018a,liao2025beyondmimic,xie2025kungfubot,hanKungfuBot2LearningVersatile2025a,ma2026extremergmt,zhang2026any2track,he2025asap,sleiman2026zest}.
Recent behavior foundation models (BFMs) further demonstrate generalization to unseen motions by scaling data diversity and model capacity~\cite{luo2025sonic,zeng2026scalebfm,qi2026humanoidgpt,chen2026holomotion}.
Such controllers support teleoperation for collecting robot demonstrations and provide a motor interface for higher-level planners.
Extending these capabilities across real-world terrains requires controllers to perceive local support geometry and adapt the commanded behavior accordingly.
Yet terrain-agnostic controllers lack both terrain observations and training supervision for this adaptation.

A key bottleneck is obtaining scalable paired supervision that links the same behavior across flat ground and different terrains.
Existing flat-ground motion datasets lack terrain-adapted counterparts, and scaling their repertoire alone does not fill this gap.
Collecting such pairs directly is difficult: terrain-specific demonstrations require additional motion capture and corresponding scene geometry to reproduce the support conditions in simulation~\cite{zhuang2026deep}, while separate recordings on flat ground and terrain do not automatically align.
These requirements make direct collection costly to scale across behaviors and terrains.
This calls for generating terrain-adapted counterparts from flat-ground motions while preserving their characteristics and adjusting posture and contacts, including for hand-supported behaviors.

To this end, we present \textbf{NEXUS}, a perceptive whole-body control framework for teleoperation under human--robot terrain mismatch.
We first develop a terrain-aware motion adaptation algorithm that generates temporally aligned motion pairs across behaviors and terrains without per-motion or per-terrain tuning, adjusting hand and foot contacts while largely preserving non-contact poses.
These pairs enable teacher--student learning: a teacher tracks the adapted references and supervises a student conditioned on flat-ground commands, proprioceptive history, and onboard depth.
The student learns to adapt commanded behaviors to local terrain without requiring adapted references at deployment.

Our contributions are threefold:
\begin{enumerate}
  \item A perceptive whole-body control framework that executes flat-ground human motion commands across terrains, combining broad behavioral coverage with terrain adaptability and tracking fidelity.
  \item A scalable, contact-guided adaptation algorithm that generates paired references without per-motion or per-terrain tuning, improving terrain contact consistency while preserving source motion characteristics.
  \item Zero-shot real-world validation of diverse whole-body behaviors on unseen terrains through real-time teleoperation, with the operator remaining on flat ground.
\end{enumerate}

\section{Related Work}
\label{sec:related-work}

\subsection{Kinematic Terrain Adaptation}

To preserve motion--environment interactions during retargeting, OmniRetarget~\cite{yang2025omniretarget} uses an interaction mesh, which also supports terrain augmentation around the original scene configuration.

Transferring a motion to substantially different terrain poses a broader challenge than augmenting the original scene.
Character animation methods address this through terrain-relative contact descriptors~\cite{cheynel2025reconform} or by adapting single-rigid-body trajectories and reconstructing whole-body motion~\cite{cao2025srbtrack}.
For humanoids, TCRS~\cite{wang2026perceptive} combines sampling-based foot-trajectory optimization with root-height adjustment and leg IK, leaving non-leg joints unchanged.
In contrast, NEXUS efficiently adapts both hand and foot contacts without per-motion or per-terrain tuning, generating high-quality, temporally aligned reference pairs at scale for policy learning.

\subsection{Perceptive Whole-Body Humanoid Control}

Perceptive locomotion traverses challenging terrain under navigation goals or velocity commands~\cite{zhu2026hiking,zhang2026rpl,wang2026more,wang2025beamdojo,he2025ame,longLearningHumanoidLocomotion2024a}, which leave the desired whole-body poses and timing unspecified.
Motion references specify the coordinated poses and timing of whole-body behaviors, enabling diverse and dynamic skill learning from kinematic demonstrations~\cite{liao2025beyondmimic,xie2025kungfubot,hanKungfuBot2LearningVersatile2025a,ma2026extremergmt,zhang2026any2track,he2025asap,sleiman2026zest,yin2026unitracker,chen2026gmt,xie2026textop,wang2025bumblebee,ji2026exbody2}.
While scaling motion data and policy capacity enables generalization to unseen motions~\cite{luo2025sonic,zeng2026scalebfm,qi2026humanoidgpt,chen2026holomotion}, these controllers remain largely terrain-agnostic, with limited ability to adapt reference behaviors to different terrain conditions.

Combining perception with motion learning enables whole-body locomotion across challenging terrain using terrain-specific skills or references~\cite{zhuang2026deep,wu2026perceptive,zhang2026wholebodylocomotion}.
PMT~\cite{wang2026perceptive} learns to adapt flat-ground motions to the robot's local terrain from terrain-adapted supervision over a locomotion- and dance-oriented corpus, but shows limited generalization to a broader repertoire of whole-body behaviors even on flat ground.
Its reference synthesis is computationally expensive and yields inconsistent adaptation quality across motions and terrains, limiting scalable data generation over diverse motion--terrain combinations.
Using a large-scale set of high-quality paired kinematic trajectories generated by our adaptation method, NEXUS trains a perceptive whole-body controller to support broader behavioral coverage and generalization to unseen terrains.

\begin{figure*}[t]
  \centering
  \includegraphics[width=\textwidth]{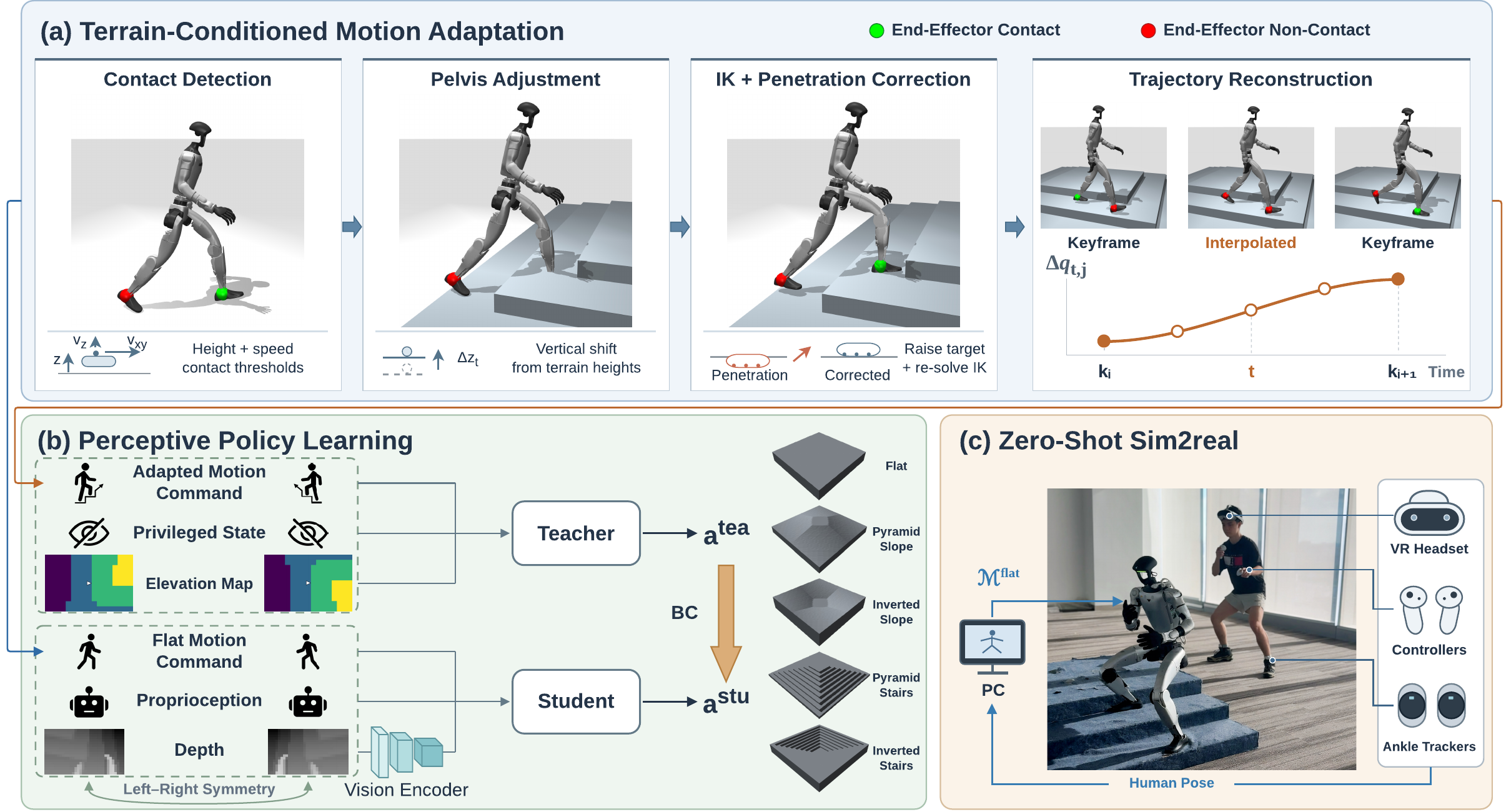}
  \caption{\textbf{NEXUS overview.} (a) Contact-guided motion adaptation: contact detection, pelvis adjustment, IK with penetration correction, and trajectory reconstruction from contact keyframes. Contact markers are illustrative. (b) A privileged teacher trained with PPO supervises a student through DAgger-style behavior cloning (BC) using paired adapted and flat-ground motion commands. (c) Zero-shot whole-body teleoperation with onboard depth sensing and policy inference.}
  \label{fig:overview}
\end{figure*}

\section{Method}
\label{sec:method}

\subsection{Overview}
\label{sec:overview}

We consider whole-body teleoperation under human--robot terrain mismatch, where a human operator moves on flat ground while a humanoid robot executes the commanded behavior over terrain.
The operator's retargeted motion defines a flat-ground reference $\mathcal M^{\mathrm{flat}}$; the robot must adjust its motion to local support geometry while preserving the commanded behavior.
We learn a policy
\begin{equation}
  \mathbf a_t=\pi\!\left(\mathbf m_t^{\mathrm{flat}},\mathbf o_t^{\mathrm{prop}},\mathbf o_t^{\mathrm{ext}}\right),
  \label{eq:teleoperation-policy}
\end{equation}
where $\mathbf m_t^{\mathrm{flat}}$ encodes $\mathcal M^{\mathrm{flat}}$, $\mathbf o_t^{\mathrm{prop}}$ denotes the robot's proprioceptive history, and $\mathbf o_t^{\mathrm{ext}}$ provides exteroceptive observations of its local terrain.
The action $\mathbf a_t$ specifies scaled offsets from nominal joint positions for PD control.
Robot-state quantities carry no reference superscript; $\mathrm{flat}$ and $\mathrm{adapt}$ denote the flat-ground and terrain-adapted references, respectively.

As shown in Fig.~\ref{fig:overview}, NEXUS first constructs terrain-adapted references $\mathcal M^{\mathrm{adapt}}$ temporally aligned with $\mathcal M^{\mathrm{flat}}$ through offline kinematic adaptation.
We train a privileged teacher $\pi^{\text{tea}}$ with PPO~\cite{schulman2017ppo} to track these references, then distill it into a perceptive student $\pi^{\text{stu}}$ using DAgger~\cite{rossReductionImitationLearning2011b}.
The student implements $\pi$ in Eq.~\eqref{eq:teleoperation-policy} using a three-layer CNN depth encoder and an MLP action head.

\subsection{Terrain-Conditioned Motion Adaptation}
\label{sec:terrain-adaptation}

Given a flat-ground robot motion $\mathcal{M}^{\mathrm{flat}}$ and a fixed terrain, we construct a reference $\mathcal{M}^{\mathrm{adapt}}$ by correcting poses at source-contact frames and propagating their corrections through time.
Each frame $t$ of $\mathcal M$ comprises floating-base position $\mathbf p_t$ (pelvis), unit-quaternion orientation, and joint angles $\mathbf q_t$.
Here, $\mathbf p_{t,b}$ denotes body $b$'s position, and $p_{t,z}$ is pelvis height, all in world coordinates.
We use a single configuration for all motions and terrains.

\textbf{Contact detection.}
We infer a contact label $c_{t,i}\in\{0,1\}$ for each ankle and wrist link from its origin's height, horizontal speed, and absolute vertical speed in $\mathcal M^{\mathrm{flat}}$, using separate contact entry and exit thresholds.
Both speed thresholds scale with the pelvis's horizontal speed after clamping that speed to at least $1\,\mathrm{m/s}$.
We designate every frame with at least one active contact as a contact keyframe, forming the set $\mathcal K$; clips with $\mathcal K=\varnothing$ are excluded.

\textbf{Terrain targets and pelvis adjustment.}
After horizontal placement on the terrain, let $\mathbf p_{t,i}^{\mathrm{flat}}=(x_{t,i},y_{t,i},z_{t,i})^\top$ denote active end effector $i$'s source position, with source ground at $z=0$.
Downward raycasts provide terrain heights $h(x,y)$ and normals; missing hits and surfaces with insufficient upward normal components are ignored.
For valid contacts, the initial position target is
\begin{equation}
  \mathbf{p}_{t,i}^{\mathrm{des}} =
  \begin{bmatrix}
    x_{t,i} & y_{t,i} & z_{t,i}+h(x_{t,i},y_{t,i})
  \end{bmatrix}^{\!\top}.
  \label{eq:terrain-contact-target}
\end{equation}
The added $z_{t,i}$ accounts for the source ankle or wrist origin's height above the ground.
Let $\mathcal C_t$ be the set of active end effectors with valid terrain queries, and let $d_{t,i}=h(x_{t,i},y_{t,i})$ be their terrain-height offsets from the source flat ground.
For $\mathcal C_t\ne\varnothing$, we shift the pelvis by the offset with the largest absolute value, retaining its sign:
\begin{equation}
  i_t^\star=\operatorname*{arg\,max}_{i\in\mathcal C_t}|d_{t,i}|,
  \qquad \Delta z_t=d_{t,i_t^\star}.
  \label{eq:terrain-root-adjustment}
\end{equation}
If $\mathcal C_t=\varnothing$, we set $\Delta z_t=0$.

\textbf{Contact-keyframe IK.}
We adjust the robot's pose at contact keyframes so that the contacting hands and feet meet the terrain surface.
For each keyframe $t\in\mathcal K$, we solve inverse kinematics (IK) with the pelvis fixed at the terrain-adjusted height.
For each active limb, standard analytic two-bone IK constructs knee or elbow targets from the source limb's lengths and bend direction, and restricts end-effector reach by bend limits.
For active end effector $i$ on limb $\ell$, let $\widehat{\mathbf p}_m$ and $\widehat{\mathbf p}_e$ be the intermediate and reach-clamped targets constructed from $\mathbf p_{t,i}^{\mathrm{des}}$.
We denote its four optimized joint angles by $\mathbf q_{\ell}$ and its source angles, taken from $\mathbf q_t^{\mathrm{flat}}$, by $\mathbf q_{\ell}^{\mathrm{flat}}$.
Suppressing $t,i$ locally and fixing the other degrees of freedom, we solve
\begin{equation}
  \begin{aligned}
    &\min_{\mathbf q_{\ell}^{\min}\leq\mathbf q_{\ell}\leq\mathbf q_{\ell}^{\max}}
    \quad \|\mathbf f_m(\mathbf q_{\ell})-\widehat{\mathbf p}_m\|_2^2 \\
    &\quad +\|\mathbf f_e(\mathbf q_{\ell})-\widehat{\mathbf p}_e\|_2^2
    +\lambda^2\|\mathbf q_{\ell}-\mathbf q_{\ell}^{\mathrm{flat}}\|_2^2,
  \end{aligned}
  \label{eq:terrain-limb-ik}
\end{equation}
where $\mathbf f_m,\mathbf f_e$ are world-position forward kinematics, $\mathbf q_{\ell}^{\min},\mathbf q_{\ell}^{\max}$ are joint limits, and $\lambda$ weights the posture residual.
The posture term regularizes the bounded least-squares solve toward the source configuration.
To orient the soles toward the local support surface, we separately optimize the ankle joints to align each contacting foot with the terrain normal, subject to joint limits and a cap on the orientation correction.
We apply this alignment only to feet, as the robot's hands are modeled as capsules without a flat support surface.

\textbf{Penetration correction.}
After fitting each contacting limb, we sample its foot or hand collision geometry and raise its position target by the largest sampled penetration, then repeat the position solve and, for feet, orientation alignment.
If the knee or elbow penetrates, we search bounded rotations of the limb's bend plane and re-solve the chain.
We accept candidates that satisfy sampled clearance and end-effector drift tolerances without increasing sampled end-effector penetration; otherwise, we restore the preceding pose.
Non-contact limbs undergo end-effector and intermediate-link penetration correction at $t\in\mathcal K$.

\textbf{Trajectory reconstruction.}
At each $k\in\mathcal K$, the pose after penetration correction, denoted by $\mathrm{IK}$, defines the source-relative correction
\begin{equation}
  \boldsymbol\delta_k=
  \begin{bmatrix}
    \mathbf q_k^{\mathrm{IK}}-\mathbf q_k^{\mathrm{flat}}\\
    p_{k,z}^{\mathrm{IK}}-p_{k,z}^{\mathrm{flat}}
  \end{bmatrix}.
  \label{eq:terrain-keyframe-correction}
\end{equation}
Within the keyframe span, we apply shape-preserving piecewise cubic Hermite interpolation (PCHIP)~\cite{fritsch1984monotone} componentwise:
\begin{equation}
  \begin{bmatrix}\Delta\mathbf q_t\\\Delta z_t\end{bmatrix}
  =\operatorname{PCHIP}\!\left(\{(k,\boldsymbol\delta_k)\}_{k\in\mathcal K}\right)(t).
  \label{eq:terrain-correction-interpolation}
\end{equation}
At $k\in\mathcal K$, $\Delta z_k$ equals the pelvis shift in Eq.~\eqref{eq:terrain-root-adjustment}. Outside the keyframe span, we hold the nearest correction constant.
Adding the corrections to $\mathcal M^{\mathrm{flat}}$ yields
\begin{equation}
  \begin{aligned}
    \mathbf q_t^{\mathrm{adapt}}
      &= \mathbf q_t^{\mathrm{flat}} + \Delta\mathbf q_t, \\
    p_{t,z}^{\mathrm{adapt}}
      &= p_{t,z}^{\mathrm{flat}} + \Delta z_t,
  \end{aligned}
  \label{eq:terrain-motion-reconstruction}
\end{equation}
The horizontally placed source root's $x,y$ coordinates and orientation remain unchanged.
Forward kinematics and finite differences recover body poses and velocities.
$\mathcal M^{\mathrm{adapt}}$ and $\mathcal M^{\mathrm{flat}}$ remain aligned frame by frame, providing paired supervision for policy learning.

\begin{table*}[t]
  \centering
  \caption{Kinematic motion adaptation quality and efficiency.}
  \label{tab:terrain-adaptation}
  \begingroup
  \fontsize{9}{10}\selectfont
  \setlength{\tabcolsep}{3pt}
  \renewcommand{\arraystretch}{1.12}
  \begin{tabular*}{\textwidth}{@{\extracolsep{\fill}}lcccccc@{}}
    \toprule
    Method & VTR (\%)\,$\uparrow$ & Penetration (cm)\,$\downarrow$
      & Floating (cm)\,$\downarrow$ & CP (\%)\,$\uparrow$ & Deviation (rad)\,$\downarrow$ & Generation RTF\,$\downarrow$ \\
    \midrule
    \rowcolor{black!8}
    \multicolumn{7}{l}{Motions without hand contact} \\
    \midrule
    \textbf{NEXUS (ours)} & \textbf{92.08} & 0.814 & 0.880 & \textbf{79.12} & \textbf{0.00984} & 1.646 \\
    TCRS (in PMT) & 89.69 & \textbf{0.606} & 2.965 & 29.63 & 0.08884 & 8.038 \\
    Root-only    & 63.37 & 2.865 & \textbf{0.817} & 54.25 & 0.00000 & \textbf{0.260} \\
    \midrule
    \rowcolor{black!8}
    \multicolumn{7}{l}{Motions with hand contact} \\
    \midrule
    \textbf{NEXUS (ours)} & \textbf{80.53} & \textbf{1.630} & \textbf{0.946} & \textbf{71.21} & \textbf{0.01473} & 1.864 \\
    TCRS (in PMT) & 67.95 & 4.432 & 4.835 & 20.61 & 0.12213 & 7.887 \\
    Root-only    & 55.69 & 3.839 & 1.230 & 48.94 & 0.00000 & \textbf{0.109} \\
    \bottomrule
  \end{tabular*}
  \par\smallskip
  \parbox{\textwidth}{\fontsize{8}{9}\selectfont
    VTR measures geometric validity over time; penetration and floating capture terrain violations; CP measures contact retention; deviation measures non-contact pose changes; Generation RTF measures computation time per second of input motion; lower values indicate faster generation.
    Root-only preserves joint angles by construction, so its zero deviation is excluded from bolding.}
  \endgroup
\end{table*}

\subsection{Perceptive Policy Learning}
\label{sec:perceptive-policy-learning}

The paired references connect the tracking objective on each terrain to the flat-ground motion commands available during teleoperation.

\textbf{Teacher observations.}
The teacher receives privileged motion features $\mathbf m_t^{\mathrm{adapt}}$ that encode reference-window root pose errors and velocities, joint states $(\mathbf q^{\mathrm{adapt}},\dot{\mathbf q}^{\mathrm{adapt}})$, projected gravity $\mathbf g^{\mathrm{adapt}}$, and root-relative body poses, including pose tracking errors.
Privileged feedback $\mathbf o_t^{\mathrm{priv}}$ contains root-relative body poses $(\mathbf p_{t,b}^{\mathrm{rel}},\mathbf R_{t,b}^{\mathrm{rel}})$ and velocities $(\mathbf v_{t,b}^{\mathrm{rel}},\boldsymbol\omega_{t,b}^{\mathrm{rel}})$, histories of base-frame linear and angular velocities $(\mathbf v^{\mathrm{base}},\boldsymbol\omega^{\mathrm{base}})$, projected gravity $\mathbf g$, and joint states $(\mathbf q,\dot{\mathbf q})$, plus current root world-height and projected-gravity errors.
It also includes the previous action $\mathbf a_{t-1}$, PD joint targets $\mathbf q_t^{\mathrm{des}}$, actuator torques $\boldsymbol\tau_t$, and actual and reference end-effector contacts.
Together with the root-relative terrain height map $\mathbf H_t\in\mathbb R^{11\times17}$, these form the grouped observation
\begin{equation}
  \mathbf o_t^{\text{tea}}=
    [\mathbf m_t^{\mathrm{adapt}},\mathbf o_t^{\mathrm{priv}},\mathbf H_t].
  \label{eq:teacher-policy}
\end{equation}

\textbf{Student observations.}
The student receives the paired flat-ground motion features $\mathbf m_t^{\mathrm{flat}}$, proprioceptive history $\mathbf o_t^{\mathrm{prop}}$, and a single depth image $\mathbf D_t$, which serves as the exteroceptive input $\mathbf o_t^{\mathrm{ext}}$ in NEXUS.
The motion window specifies root displacement and orientation, absolute root height $p_z^{\mathrm{flat}}$, joint positions $\mathbf q^{\mathrm{flat}}$, and projected gravity $\mathbf g^{\mathrm{flat}}$.
Displacements are expressed in the window's first root frame; orientations are relative to the robot's current root.
At each time step $t$, we stack base-frame angular velocity $\boldsymbol\omega_t^{\mathrm{base}}$, projected gravity $\mathbf g_t$, joint offsets $\mathbf q_t-\mathbf q^0$ from nominal angles $\mathbf q^0$, joint velocities $\dot{\mathbf q}_t$, and the previous action $\mathbf a_{t-1}$.
The grouped student observation is
\begin{equation}
  \mathbf o_t^{\text{stu}}=
    [\mathbf m_t^{\mathrm{flat}},\mathbf o_t^{\mathrm{prop}},\mathbf D_t].
  \label{eq:student-policy}
\end{equation}

\textbf{Left--right symmetry.}
Sagittal reflection swaps left/right joints, bodies, and contacts.
Joint states, actions, PD targets, and torques follow signed joint permutations.
Positions, linear velocities, and gravity negate the lateral component; angular velocities negate forward and vertical components.
Rotations follow $\mathbf R\mapsto\mathbf S\mathbf R\mathbf S$, $\mathbf S=\operatorname{diag}(1,-1,1)$.
Reference and history frames retain temporal order.
Elevation maps reverse lateral grid order; depth images flip horizontally, preserving scalar values.
We train the teacher with PPO and symmetry regularization:
\begin{equation}
 \mathcal L_{\mathrm{sym}}^{\text{tea}}=
 \mathbb E\!\left[\operatorname{MSE}\!\left(
 \pi^{\text{tea}}(\mathsf M_o\mathbf o^{\text{tea}}),
 \operatorname{sg}[\mathsf M_a\pi^{\text{tea}}(\mathbf o^{\text{tea}})]
 \right)\right],
 \label{eq:teacher-symmetry}
\end{equation}
Here, $\mathsf M_o$ and $\mathsf M_a$ reflect observations and actions, respectively; policy outputs denote mean actions, and $\operatorname{sg}$ stops gradients. The expectation is over teacher rollout observations.
BC re-queries the frozen teacher on the augmented batch:
\begin{equation}
 \mathcal L_{\mathrm{aug}}^{\text{stu}}=
 \mathbb E_{\tilde{\mathbf o}}\!\left[\operatorname{MSE}\!\left(
 \pi^{\text{stu}}(\tilde{\mathbf o}^{\text{stu}}),
 \pi^{\text{tea}}(\tilde{\mathbf o}^{\text{tea}})
 \right)\right],
 \label{eq:student-symmetry}
\end{equation}
Here, $\tilde{\mathbf o}$ denotes paired teacher/student observations at student-visited states, including original and mirrored samples.

\textbf{Rewards, termination, and adaptive sampling.}
For teacher training, the total reward combines motion tracking, contact objectives, control regularization, and survival:
\begin{equation}
  r_t=
    r_t^{\mathrm{track}}+r_t^{\mathrm{contact}}
    +r_t^{\mathrm{reg}}+r_t^{\mathrm{alive}},
  \label{eq:total-reward}
\end{equation}
where each component is the weighted reward contribution per control step.
The tracking component $r_t^{\mathrm{track}}$ matches root pose and velocity, root-relative body poses, global body velocities, and joint states to the adapted reference.
The contact component $r_t^{\mathrm{contact}}$ penalizes foot and hand contact-timing and terrain-clearance discrepancies, as well as insufficient foot support.
The regularization component $r_t^{\mathrm{reg}}$ penalizes action changes, joint velocities, joint-position and torque-limit violations, and self-collisions; $r_t^{\mathrm{alive}}$ rewards survival.

For termination, $b$ indexes the ankle and wrist links.
The local ground heights are $h_t=h(p_{t,x},p_{t,y})$ and $h_t^{\mathrm{adapt}}=h(p_{t,x}^{\mathrm{adapt}},p_{t,y}^{\mathrm{adapt}})$.
Termination uses root clearance and root-relative end-effector heights, making the criteria insensitive to horizontal drift alone:
\begin{equation}
  \begin{aligned}
    e_t^{\mathrm{clear}}
      &= \big|(p_{t,z}-h_t)
         -(p_{t,z}^{\mathrm{adapt}}-h_t^{\mathrm{adapt}})\big|, \\[2pt]
    e_{t,b}^{\mathrm{rel}}
      &= \big|(p_{t,b,z}-p_{t,z})
         -(p_{t,b,z}^{\mathrm{adapt}}-p_{t,z}^{\mathrm{adapt}})\big|.
  \end{aligned}
  \label{eq:terrain-termination}
\end{equation}
Episodes end if $e_t^{\mathrm{clear}}$ or any $e_{t,b}^{\mathrm{rel}}$ exceeds $0.25\,\mathrm{m}$, if the absolute difference between the $z$-components of projected gravity in the robot and reference root frames exceeds $0.8$, or when the time limit is reached.

Similar to~\cite{luo2025sonic}, we adopt adaptive sampling that prioritizes terrain--motion--phase segments with higher failure rates while retaining a uniform sampling component.

\begin{figure*}[t]
  \centering
  \includegraphics[width=\textwidth]{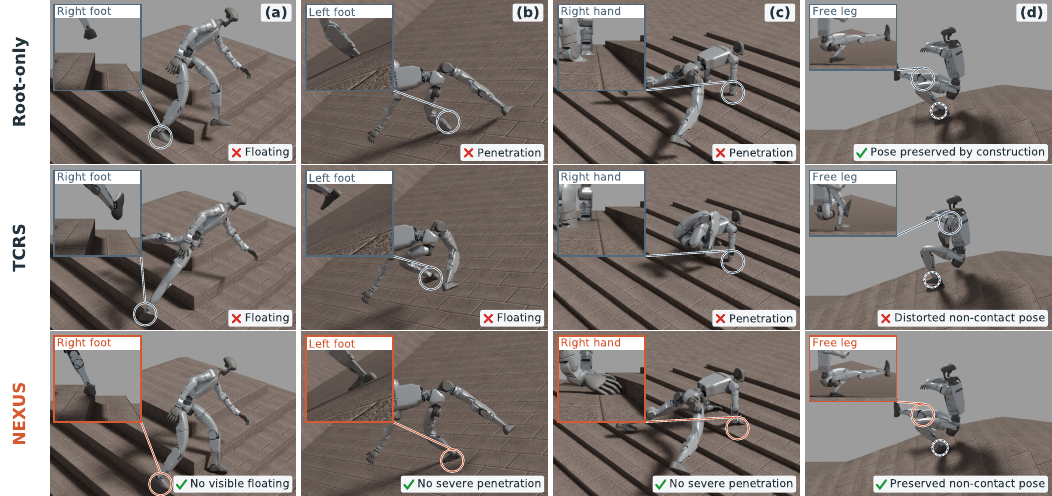}
  \caption{\textbf{Baseline failures and NEXUS improvements at matching frames:} (a) foot floating in both baselines; (b) foot penetration in Root-only and floating in TCRS; (c) hand penetration in both baselines; (d) TCRS distorts the non-contact pose by retracting the extended free leg. NEXUS improves contact while preserving the leg's extension. Insets magnify circled regions; dashed circles mark the support foot.}
  \label{fig:terrain-adaptation-qualitative}
\end{figure*}

\section{Experiments}
\label{sec:experiments}

Our experiments evaluate: (1) the scalability, geometric quality, and motion fidelity of the NEXUS motion adaptation pipeline; (2) the behavioral coverage, terrain adaptability, and tracking fidelity of our perceptive policy; (3) the contributions of vision and teacher--student training to that policy; and (4) the NEXUS framework's zero-shot transfer to real-world whole-body teleoperation.

\indent\textbf{Training setup.}
We train our policies in mjlab~\cite{zakka2026mjlab} using four NVIDIA H100 GPUs, with 16,384 parallel environments per GPU.
The student policy receives a 10-frame history of proprioceptive observations and a single depth image.
Training terrains comprise flat ground and procedurally generated ascending and descending slopes with gradient $0.30$ (rise/run), and stairs with $0.10$\,m step height and $0.30$\,m tread depth.
We use curated subsets of flat-ground motions from SEED~\cite{bonesSeed}, LaFAN1~\cite{harvey2020robust}, AMASS~\cite{mahmood2019amass}, and 100STYLE~\cite{mason2022local}, totaling nearly 1,000 hours including terrain-adapted variants.

\indent\textbf{Metrics.}
For kinematic motion evaluation, valid time ratio (VTR) is the fraction of frames with body penetration $\leq5$\,cm and expected-contact gap $\leq5$\,cm and penetration $\leq2$\,cm; only VTR excludes source contact events shorter than $0.1$\,s.
Contact preservation (CP) is the fraction of expected contacts with gap $\leq2$\,cm and penetration $\leq0.5$\,cm.
Penetration measures the mean framewise maximum body penetration; floating measures the mean positive gap at expected contacts, both including zeros.
Deviation is the mean absolute change from source angles in non-contact limb joints and all waist joints.
Generation real-time factor (RTF, dimensionless) divides launch-to-exit wall time by input motion duration, including preparation, adaptation, and saving but excluding common evaluation and reporting.

For policy evaluation, success rate (SR) is the fraction of trials that complete the entire reference motion.
Mean per-keypoint position/rotation errors (MPKPE/MPKRE) measure local body tracking errors.
Root position error (Root Pos) measures root displacement error after initial pose alignment.
Contact F1 measures agreement with reference contact labels.
Tracking errors and contact F1 use each policy's own pre-failure window.

\subsection{Kinematic Motion Adaptation}
\label{sec:exp-terrain-adaptation}

We compare NEXUS with TCRS~\cite{wang2026perceptive} and Root-only, a simple baseline that adds the local terrain height to the source root height, retains all source joint angles, and reconstructs body poses through forward kinematics (FK).
The benchmark comprises 55 full-length G1 motions (2.49 hours) from LaFAN1~\cite{harvey2020robust} and OmniXtreme~\cite{wang2026omnixtreme}, grouped by the presence of hand contacts, and 16 terrains covering four settings each of ascending/descending stairs and slopes.
Each method uses one fixed configuration across all motion--terrain pairs.
Quality metrics are averaged over seeds and terrains, with equal weight per motion within each group.

Table~\ref{tab:terrain-adaptation} evaluates whether the adapted motions satisfy terrain contact constraints while preserving the source pose.

\indent\textbf{Scalability.}
NEXUS processes all 880 motion--terrain pairs, generating 39.87 hours of adapted motion, including hand-contact motions, without per-motion or per-terrain tuning.
Given TCRS's high computational cost, timing uses a prespecified eight-motion subset balanced by source and contact group across all 16 terrains, with one run per method--case pair.
Each process uses one CPU core and one thread on an Intel Xeon Gold 6530, with 32 concurrent workers after interference calibration.
Paired timing includes valid complete returns, including TCRS's rejected candidates; its six process errors leave 63/64 locomotion and 59/64 hand-contact cases, while NEXUS and Root-only each complete 128/128 attempts.
NEXUS reduces mean generation RTF relative to TCRS by factors of 4.88 and 4.23 in the two groups, respectively.
Root-only is fastest, but its geometric limitations below show that generation cost alone does not establish useful terrain adaptation.

\indent\textbf{Geometric quality.}
NEXUS achieves the highest contact preservation and valid time ratio in both motion groups (Table~\ref{tab:terrain-adaptation}).
As Fig.~\ref{fig:terrain-adaptation-qualitative}(a,b) illustrates, reducing penetration alone can leave a support foot floating, so both errors must be assessed together.
The improvements also extend to hand contacts: NEXUS attains the lowest penetration and floating in this group and brings the palm near the surface where both baselines place it inside the terrain (Fig.~\ref{fig:terrain-adaptation-qualitative}(c)).

\indent\textbf{Motion fidelity.}
Terrain adaptation should preserve source poses in non-contact parts of the body while adjusting the supporting limbs.
NEXUS achieves this with less deviation in non-contact joints than TCRS while improving contact consistency.
In Fig.~\ref{fig:terrain-adaptation-qualitative}(d), TCRS corrects foot contact but distorts the free-leg pose, whereas NEXUS preserves its extension.
We do not train controllers on these baseline references: Root-only penetration can compromise reference-based initialization, while TCRS exhibits occasional pose distortions, generation errors, and higher computational cost.

\begin{figure*}[t]
  \centering
  \includegraphics[width=\textwidth]{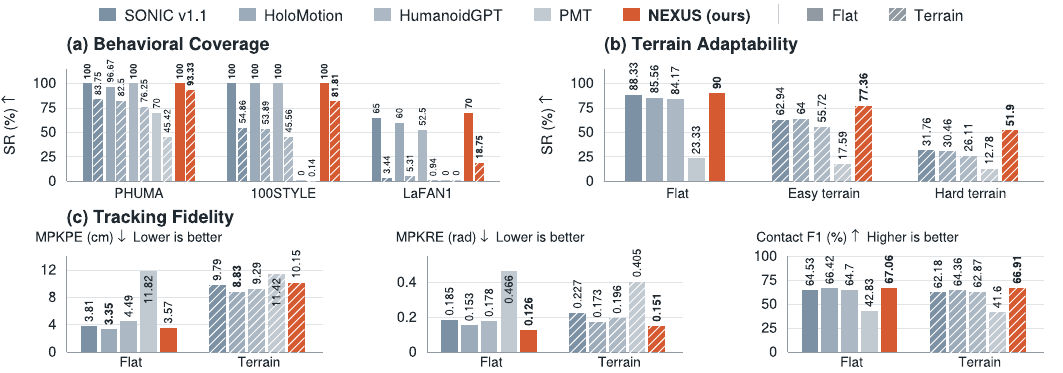}
  \caption{\textbf{Whole-body control generalization compared with four open-source controllers.}
  (a) NEXUS achieves the highest motion completion rates across the three dataset subsets, including ties.
  (b) NEXUS maintains the highest completion rates across terrain difficulty levels.
  (c) NEXUS attains the lowest body orientation error and highest foot-contact F1 on both flat ground and terrain.
  Tracking metrics use each controller's own pre-failure window.}
  \label{fig:tracking-comparison}
\end{figure*}
\subsection{Whole-Body Control Generalization}
\label{sec:exp-motion-tracking}

We compare NEXUS with public checkpoints of SONIC v1.1~\cite{luo2025sonic}, HumanoidGPT~\cite{qi2026humanoidgpt}, HoloMotion~\cite{chen2026holomotion}, and PMT~\cite{wang2026perceptive}.
We evaluate all 160 motions (2.61~h) in three fixed public evaluation subsets~\cite{sim2realTrackingEvaluation} drawn from PHUMA~\cite{lee2025phuma}, 100STYLE, and LaFAN1 (Fig.~\ref{fig:tracking-comparison}(a)).
These subsets comprise short clips with mostly limited translation, directional locomotion, and substantially longer sequences covering diverse whole-body behaviors, respectively.
We test each checkpoint on flat ground and ascending and descending stairs and slopes.
The easy terrain level uses 5\,cm steps and slopes of $8.5^\circ$; the hard level uses 15\,cm steps and slopes of $24.2^\circ$, with 30\,cm stair treads at both levels.
NEXUS trains on 10\,cm steps and $16.7^\circ$ slopes, differing from both test levels.
Milder terrain tests blind controllers' tolerance to support-height changes; harder terrain requires larger departures from flat-ground execution.
Checkpoints remain fixed across conditions and receive identical flat-ground commands; adapted trajectories provide pose-tracking references.
We average motions within each subset, then weight subsets and terrain conditions equally.
On both flat and non-flat terrains, we evaluate foot-contact F1 using threshold-based labels from the original flat-ground references and robot trajectories.

\indent\textbf{Behavioral coverage.}
Flat-ground evaluation separates behavioral generalization from terrain adaptation (Fig.~\ref{fig:tracking-comparison}(a)).
NEXUS and SONIC track a broad range of motions, whereas PMT fails on the 100STYLE and LaFAN1 subsets even on flat ground.
PMT's terrain-adapted supervision therefore does not by itself ensure broad behavioral coverage; its locomotion- and dance-oriented corpus offers a plausible explanation for this limitation~\cite{wang2026perceptive}.
The comparison suggests that diversity in the underlying motion corpus remains important alongside terrain adaptation.

\indent\textbf{Terrain adaptability.}
Terrain-agnostic controllers retain some success on mild terrain but deteriorate as support geometry departs further from flat ground (Fig.~\ref{fig:tracking-comparison}(b)).
Their proprioceptive feedback can accommodate modest disturbances, but their training does not explicitly teach how flat-ground commands should change with local terrain geometry.
Even motions tracked reliably on flat ground become difficult under changed support conditions.
NEXUS's stronger performance across terrain levels supports the value of pairing flat-ground commands with terrain-adapted supervision and robot-side perception.

\indent\textbf{Tracking fidelity.}
NEXUS combines the lowest body orientation error with the highest foot-contact F1 on both flat ground and terrain (Fig.~\ref{fig:tracking-comparison}(c)), indicating that its broader completion capability also retains orientation and contact fidelity.
HoloMotion achieves lower terrain body position error, but errors are averaged over each controller's own pre-failure window.
A controller that terminates early omits later motion segments, so a lower error alone does not establish better tracking over the complete motion; fidelity and SR must be assessed together.

\Needspace{6\baselineskip}
\subsection{Ablation Study}
\label{sec:exp-ablation}

We train all variants on the same split and test 415 held-out locomotion clips on four terrain types at five levels: 8,300 trials (20.01~h) per policy.
Steps range from 5 to 15\,cm and slope gradients from 0.15 to 0.45.
Vision and no-vision students share a teacher and training settings; single-stage PPO uses a different budget due to GPU memory constraints but is trained to convergence.

\begin{table}[t]
  \centering
  \caption{Vision and teacher--student ablations.}
  \label{tab:component-ablations}
  \begingroup
  \fontsize{8.5}{9.5}\selectfont
  \setlength{\tabcolsep}{1pt}
  \renewcommand{\arraystretch}{1.12}
  \begin{tabular*}{\columnwidth}{@{\extracolsep{\fill}}lr@{\,$\pm$\,}lr@{\,$\pm$\,}lr@{\,$\pm$\,}l@{}}
    \toprule
    Metric & \multicolumn{2}{c}{\textbf{NEXUS}} & \multicolumn{2}{c}{w/o vision} & \multicolumn{2}{c}{w/o T--S} \\
    \midrule
    SR (\%) $\uparrow$ & $\mathbf{91.072}$ & $\mathbf{14.959}$ & $87.771$ & $19.362$ & $90.964$ & $15.938$ \\
    Root Pos (cm) $\downarrow$ & $\mathbf{19.596}$ & $\mathbf{\phantom{0}5.398}$ & $20.089$ & $\phantom{0}6.651$ & $31.951$ & $\phantom{0}4.472$ \\
    MPKPE (cm) $\downarrow$ & $\mathbf{7.066}$ & $\mathbf{\phantom{0}2.949}$ & $7.603$ & $\phantom{0}3.753$ & $9.482$ & $\phantom{0}3.533$ \\
    MPKRE (rad) $\downarrow$ & $\mathbf{0.127}$ & $\mathbf{\phantom{0}0.010}$ & $0.128$ & $\phantom{0}0.010$ & $0.161$ & $\phantom{0}0.015$ \\
    Contact F1 (\%) $\uparrow$ & $\mathbf{80.662}$ & $\mathbf{\phantom{0}2.545}$ & $80.523$ & $\phantom{0}2.876$ & $79.637$ & $\phantom{0}2.859$ \\
    \bottomrule
  \end{tabular*}
  \par\vspace{1pt}
  \parbox{\columnwidth}{\raggedright\fontsize{8}{9}\selectfont
    Mean $\pm$ SD across five terrain-level means; motions and terrain types are equally weighted. Errors/F1 use adapted references and policy-specific pre-termination windows. T--S: teacher--student.}
  \endgroup
\end{table}
Table~\ref{tab:component-ablations} shows that vision improves global position tracking and completion, with a larger SR gap on the two harder terrain levels (78.10\% vs.\ 70.51\%).
Teacher--student training substantially improves tracking accuracy.
Root and body position rankings remain unchanged on a shared pre-termination window.

\subsection{Zero-Shot Real-World Deployment}
\label{sec:exp-real-world}

We deploy NEXUS on the Unitree G1 without any fine-tuning, using its onboard Intel RealSense D435i camera for depth perception.
The depth stream is configured at 60\,Hz with a resolution of $480\times270$ pixels.
We crop two pixels from each image border and resize the depth image to $32\times18$ pixels for the policy.
Depth values are clipped to 0--2\,m and normalized to $[0,1]$.
The policy runs onboard on an NVIDIA Jetson Orin NX using ONNX Runtime, with the control loop configured at 50\,Hz.
An independent onboard process acquires and preprocesses depth images, publishing them through a local ZeroMQ stream.
Human poses are retargeted on a separate PC with GMR~\cite{araujo2025retargeting}, and the resulting motion references are transmitted asynchronously to the robot via ZeroMQ.
The window uses one current, six historical, and six lookahead frames (120\,ms ahead of the current reference frame), repeating the latest received pose when incoming frames are delayed or missing.
The policy updates its motion and depth inputs without blocking on either stream and uses onboard proprioceptive feedback to produce joint-position targets for PD control.

Fig.~\ref{fig:teaser} shows NEXUS performing whole-body teleoperation across indoor and outdoor terrains.
Although training terrains contain only 10\,cm steps and slopes of approximately $16.7^\circ$, the robot continuously ascends staircases with 15\,cm steps and steps down from a 60\,cm-high platform (Fig.~\ref{fig:teaser}(a)), and can also jump down from that height.
Cartwheels are performed on indoor terrain composed of ramps and stairs (Fig.~\ref{fig:teaser}(b)) and on outdoor grassy slopes.
Outdoor deployments include staircases with approximately 15\,cm steps and grassy slopes reaching approximately $30^\circ$ (Fig.~\ref{fig:teaser}(c)).
NEXUS supports kneeling on one or both knees, crouching and squatting, single-leg behaviors such as kicking and balancing, and dynamic maneuvers such as jumps and cartwheels.
In our trials, no failures were observed during routine pose tracking; falls occurred during some drops from the 60\,cm-high platform and cartwheel attempts.
See the supplementary video for additional demonstrations.

\section{Conclusion}
\label{sec:conclusion}

We presented NEXUS, a perceptive whole-body control framework for teleoperation under human--robot terrain mismatch.
NEXUS addresses the lack of terrain-adapted supervision by efficiently generating paired motions without per-motion or per-terrain tuning.
Kinematic experiments demonstrate that our adaptation algorithm generates high-quality terrain-adapted motions at scale with low computational cost.
Through teacher--student learning, the controller learns to adapt flat-ground motion commands to local terrain using onboard perception.
Policy evaluations demonstrate broad behavioral coverage, terrain adaptability, and tracking fidelity, while zero-shot deployment enables real-time teleoperation on diverse unseen terrains.
While our adaptation algorithm handles diverse motions and terrains, it does not support overhanging terrain.
Future work could broaden its geometric coverage and connect the perceptive controller to task-level planners, with terrain-adaptive teleoperation providing a potential route to collecting demonstrations for learning autonomous behaviors.

\section*{Acknowledgments}
This work is supported by the National Natural Science Foundation of China (Grant No. 62676327, 62306242).
The SJTU team is supported by National Natural Science Foundation of China (62322603) and Shanghai Municipal Science and Technology Major Project (2025SHZDZX025D08).
We thank Chenyun Zhang, Xinzhe Liu, Yiran Wang, and Dewei Wang for their assistance with real-robot deployment, and Zibo Zhou, Yueru Chen, Bin Li, and Yunfei Ge for their helpful feedback.

\bibliographystyle{IEEEtran}
\bibliography{references}

\appendix
\makeatletter
\setlength{\@fptop}{0pt}
\setlength{\@dblfptop}{0pt}
\makeatother
\label{sec:training-appendix}

\subsection{Terrain-Conditioned Motion Adaptation Details}
\label{sec:appendix-adaptation}

Table~\ref{tab:adaptation-configuration} lists the parameters for contact detection, terrain queries, and pose correction.

\begin{figure*}[!t]
\centering
\includegraphics[width=\textwidth]{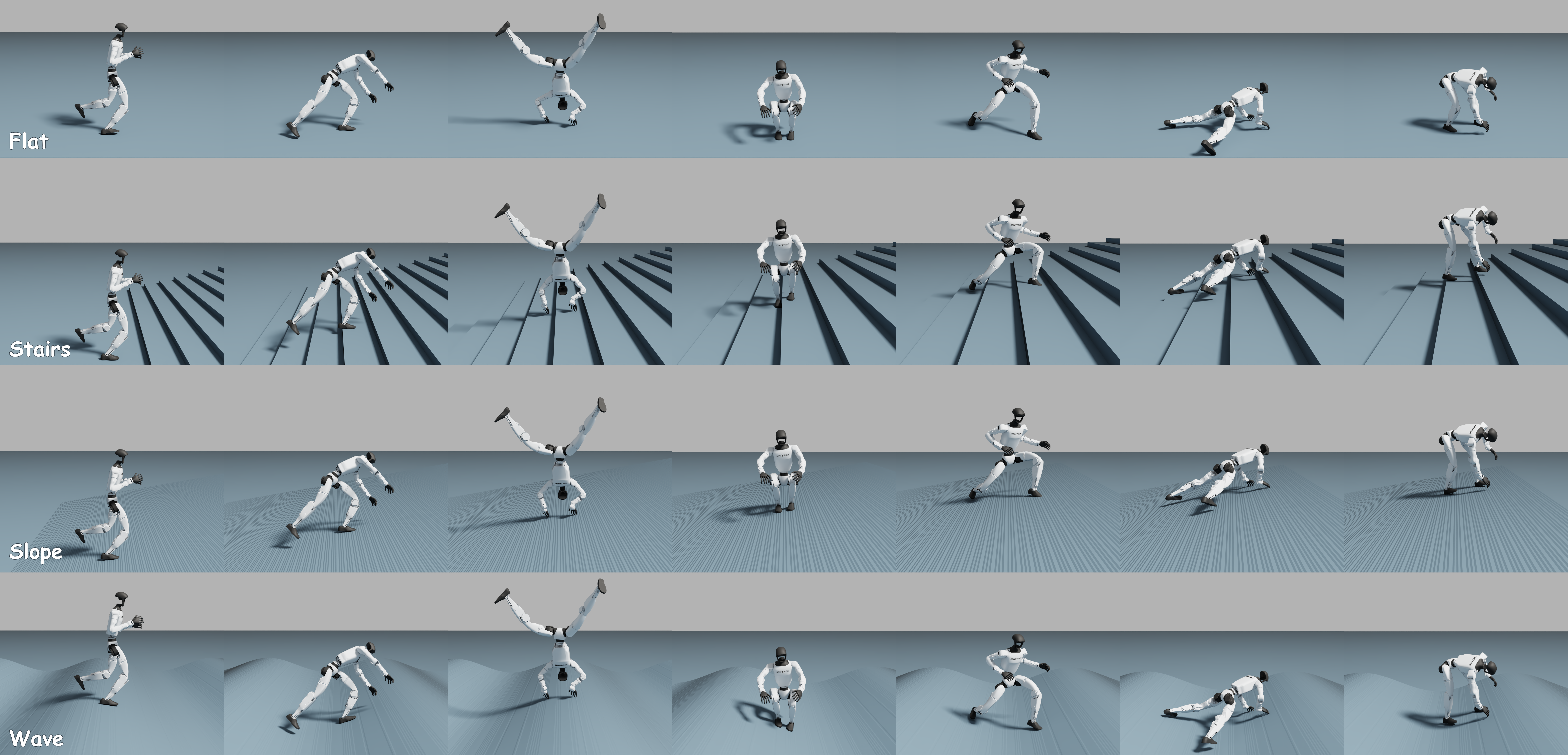}
\caption{\textbf{Motion adaptation across terrains.} Selected frames from a composite motion sequence assembled from multiple clips. Rows show the flat-ground source and its kinematic adaptations to stairs, a slope, and wavy terrain. Columns show time-aligned poses across terrains.}
\label{fig:motion-terrain-sequence}
\end{figure*}

Fig.~\ref{fig:motion-terrain-sequence} illustrates terrain-conditioned adaptation of a composite motion sequence spanning locomotion, a cartwheel, a squat, a lunge, and hand-supported transitions.

\begin{figure}[!t]
\centering
\includegraphics[width=\columnwidth]{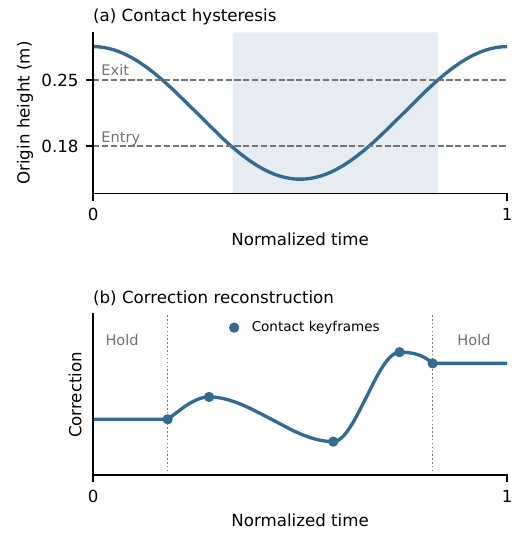}
\caption{Motion adaptation schematics. (a) Height-based contact hysteresis, assuming both speed conditions remain satisfied. Shading denotes active contact. (b) PCHIP reconstruction of a scalar correction from contact keyframes, with constant extension beyond their span. Curves are illustrative, not measured results.}
\label{fig:adaptation-details}
\end{figure}

\textbf{Contact hysteresis.}
Contact entry requires the link-origin height, horizontal speed, and absolute vertical speed to all fall below their entry thresholds. Once active, contact persists until any quantity exceeds its exit threshold. States are initialized as inactive. The speed thresholds scale with $\max(\|\mathbf v_{t,xy}^{\mathrm{flat}}\|,1\,\mathrm{m/s})$, where $\mathbf v_t^{\mathrm{flat}}$ is the source root velocity. Fig.~\ref{fig:adaptation-details}(a) illustrates the hysteresis rule.

\textbf{Terrain queries.}
We query terrain heights and surface normals using MuJoCo's raycasting function \texttt{mj\_ray}. At each queried horizontal position, a ray is cast vertically downward from above the terrain's maximum height, with robot geometries excluded. The first terrain intersection determines the height and local surface normal.

\textbf{Bounded penetration correction.}
The mid-link correction searches both signs of bend-plane rotation, with 12 evenly spaced nonzero magnitudes up to $45^\circ$. An analytic two-bone check first retains candidates that raise the intermediate target enough to compensate for its measured penetration. At most three candidates undergo the bounded joint-angle solve. Acceptance uses the resulting collision-geometry clearance, end-effector drift, and end-effector penetration, rather than the intermediate target alone. Contacting limbs have a tighter drift tolerance than non-contact limbs. If no candidate passes, the preceding pose is restored; this local procedure does not guarantee a collision-free trajectory.

\textbf{Reconstructing corrections.}
Interpolating source-relative corrections preserves the source motion for joints whose corrections remain zero at neighboring keyframes. Non-contact intervals receive corrections interpolated from surrounding contact frames. Fig.~\ref{fig:adaptation-details}(b) illustrates this reconstruction and the constant extension outside the keyframe span for one scalar correction.

\begin{table}[!t]
\centering
\caption{Motion adaptation parameters.}
\label{tab:adaptation-configuration}
\begingroup
\fontsize{8}{9.5}\selectfont
\setlength{\tabcolsep}{3pt}
\renewcommand{\arraystretch}{1.12}
\begin{tabular*}{\columnwidth}{@{\extracolsep{\fill}}lr@{}}
\toprule
Parameter & Value \\
\midrule
\rowcolor{black!8}\multicolumn{2}{l}{Contact detection: entry / exit} \\
\midrule
Link-origin height [m] & $0.18\;/\;0.25$ \\
Horizontal speed multiplier & $0.20\;/\;0.30$ \\
Absolute vertical speed multiplier & $0.15\;/\;0.25$ \\
Root horizontal speed floor [m/s] & $1.0$ \\
\midrule
\rowcolor{black!8}\multicolumn{2}{l}{Terrain queries and IK} \\
\midrule
Minimum upward normal: contact targets & $0.18$ \\
Minimum upward normal: foot alignment & $0.35$ \\
Two-bone bend limits & $2^\circ$--$145^\circ$ \\
Posture residual weight $\lambda$ & $0.08$ \\
Foot orientation correction cap & $40^\circ$ \\
\midrule
\rowcolor{black!8}\multicolumn{2}{l}{Penetration correction} \\
\midrule
Additional end-effector / mid-link clearance [m] & $0\;/\;0$ \\
Mid-link penetration tolerance [m] & $0.005$ \\
Bend-plane rotation cap & $45^\circ$ \\
Rotation magnitudes per direction & $12$ \\
Maximum candidate re-solves & $3$ \\
End-effector drift: contact / non-contact [m] & $0.005\;/\;0.05$ \\
\bottomrule
\end{tabular*}
\endgroup
\end{table}

\subsection{Policy Training Details}
\label{sec:appendix-training}

\textbf{Rewards.}
Table~\ref{tab:reward-configuration} lists the reward expressions, weights, and error scales.
\begin{table}[!t]
\centering
\caption{Reward function details.}
\label{tab:reward-configuration}
\begingroup
\fontsize{8}{9.5}\selectfont
\setlength{\tabcolsep}{3pt}
\renewcommand{\arraystretch}{1.12}
\begin{tabular*}{\columnwidth}{@{}>{\raggedright\arraybackslash\hyphenpenalty=10000}p{0.34\columnwidth}@{\hspace{4pt}}>{\centering\arraybackslash}p{0.49\columnwidth}@{\extracolsep{\fill}}r@{}}
\toprule
Reward term & Expression & Weight \\
\midrule
\rowcolor{black!8}
\multicolumn{3}{l}{Tracking} \\
\midrule
Root position & $K_{0.3}(\|\Delta\mathbf p\|^2)$ & $0.5$ \\
Root orientation & $K_{0.4}(d_R(R,R^{\mathrm{adapt}})^2)$ & $0.5$ \\
Root linear velocity & $K_{1.0}(\|\Delta\mathbf v\|^2)$ & $1.0$ \\
Root angular velocity & $K_{3.0}(\|\Delta\boldsymbol\omega\|^2)$ & $1.0$ \\
Body position & $K_{0.3}(\langle\|\Delta\mathbf p_b^{\mathrm{rel}}\|^2\rangle_b)$ & $2.0$ \\
Body orientation & $K_{0.4}(\langle d_R(R_b,R_b^{\mathrm{adapt,rel}})^2\rangle_b)$ & $2.0$ \\
Body linear velocity & $K_{1.0}(\langle\|\Delta\mathbf v_b\|^2\rangle_b)$ & $1.0$ \\
Body angular velocity & $K_{3.0}(\langle\|\Delta\boldsymbol\omega_b\|^2\rangle_b)$ & $1.0$ \\
Joint position & $K_{0.5}(\langle(\Delta q_j)^2\rangle_j)$ & $1.0$ \\
Joint velocity & $K_{3.0}(\langle(\Delta\dot q_j)^2\rangle_j)$ & $0.5$ \\
\midrule
\rowcolor{black!8}
\multicolumn{3}{l}{Contact} \\
\midrule
Contact timing & $\sum_{i\in\mathcal E} \ell_{t,i}\min(u_{t,i}-0.5,0)$ & $5.0$ \\
Contact and clearance & $\langle \chi_{t,i} K_{\sigma_h}((\Delta h_i)^2)-1\rangle_{i\in\mathcal E}$ & $1.0$ \\
Foot support & $\frac{\sum_s c_{t,f(s)}^{\mathrm{meas}}\mathbf 1[h_s>0.03]}{\max(1,\sum_s c_{t,f(s)}^{\mathrm{meas}})}$ & $-1.0$ \\
\midrule
\rowcolor{black!8}
\multicolumn{3}{l}{Regularization} \\
\midrule
Action-rate regularizer & $\|\mathbf a_t-\mathbf a_{t-1}\|^2$ & $-0.02$ \\
Joint position limits & $\sum_j([q_j^{\min}\!-q_j]_+\!+[q_j\!-q_j^{\max}]_+)$ & $-10.0$ \\
Self-collision & $N_{\mathrm{hit}}$ & $-10.0$ \\
Joint velocity & $\sum_j\dot q_j^2$ & $-5\!\times\!10^{-4}$ \\
Soft torque limits & $\sum_j[|\tau_j|/(0.75\tau_j^{\max})-1]_+$ & $-0.01$ \\
\midrule
\rowcolor{black!8}
\multicolumn{3}{l}{Alive} \\
\midrule
Survival & $\mathbf 1[\text{not terminated}]$ & $3.0$ \\
\bottomrule
\end{tabular*}
\par\smallskip
\parbox{\columnwidth}{\fontsize{8}{9}\selectfont
$K_\sigma(E)=e^{-E/\sigma^2}$; $[x]_+=\max(x,0)$; $\langle\cdot\rangle$ denotes averaging. $\Delta$ denotes robot--reference differences; $d_R$ is the rotation angle error. Indices $b,j$ denote tracked bodies and joints. Scales use m, rad, and s. Contact weights apply separately to feet and hands, with $\sigma_h=0.05/0.10$\,m, respectively. Contact and collision quantities are defined in Sec.~\ref{sec:appendix-training}.}
\endgroup
\end{table}

\textbf{Tracking and regularization.}
Body pose targets align the adapted reference to the robot root's horizontal position and heading; body velocities are compared in world coordinates. Joint-position bounds are soft limits, and $\tau_j^{\max}$ is the actuator force-limit magnitude. $N_{\mathrm{hit}}$ counts sensor-history substeps with any self-contact force above 10\,N. The teacher curriculum linearly scales all non-tracking weights from $0.5$ to $1.0$ times their listed values over the configured training horizon.

\textbf{Contact terms.}
For each end-effector group $\mathcal E$ (feet or hands), $c_{t,i}^{\mathrm{meas}}$ denotes measured contact and $\chi_{t,i}=\mathbf 1[c_{t,i}^{\mathrm{meas}}=c_{t,i}^{\mathrm{adapt}}]$. Here, $h_i$ is link-origin clearance above local terrain and $\Delta h_i=h_i-h_i^{\mathrm{adapt}}$.
The timing term is evaluated at first contact, $\ell_{t,i}=c_{t,i}^{\mathrm{meas}}(1-c_{t-1,i}^{\mathrm{meas}})$. Starting from zero, its timer gains one control step when contact states agree and loses one when they disagree:
\begin{equation}
\begin{aligned}
u_{t,i}&=\bar u_{t-1,i}+\Delta t(2\chi_{t,i}-1),\\
\bar u_{t,i}&=u_{t,i}(1-c_{t,i}^{\mathrm{meas}}).
\end{aligned}
\end{equation}
The second update resets the timer after reward evaluation whenever contact is active.
Foot support measures the fraction of sole samples more than 3\,cm above terrain among contacting feet. Samples $s$ are spaced 2\,cm apart; $h_s$ is their terrain clearance and $f(s)$ identifies their foot.

\textbf{Domain randomization.}
Table~\ref{tab:training-configuration} lists the perturbations applied at startup and reset, through periodic pushes, and in student observations.
\begin{table}[!t]
\centering
\caption{Domain randomization applied during training.}
\label{tab:training-configuration}
\begingroup
\fontsize{8}{9.5}\selectfont
\setlength{\tabcolsep}{3pt}
\renewcommand{\arraystretch}{1.12}
\begin{tabular*}{\columnwidth}{@{\extracolsep{\fill}}lr@{}}
\toprule
Randomized quantity & Distribution \\
\midrule
\rowcolor{black!8}
\multicolumn{2}{l}{Startup: teacher and student} \\
\midrule
Foot friction coefficient & $\mathcal U(0.3,1.2)$ \\
Torso CoM offset $x$ [m] & $\mathcal U(-0.025,0.025)$ \\
Torso CoM offset $y,z$ [m] & $\mathcal U(-0.05,0.05)$ \\
Joint encoder bias [rad] & $\mathcal U(-0.01,0.01)$ \\
\midrule
\rowcolor{black!8}
\multicolumn{2}{l}{Reset and pushes: teacher and student} \\
\midrule
Pelvis position offset $x,y$ [m] & $\mathcal U(-0.05,0.05)$ \\
Pelvis position offset $z$ [m] & $\mathcal U(-0.01,0.01)$ \\
Pelvis roll/pitch offset [rad] & $\mathcal U(-0.1,0.1)$ \\
Pelvis yaw offset [rad] & $\mathcal U(-0.2,0.2)$ \\
Joint position offset [rad] & $\mathcal U(-0.1,0.1)$ \\
Push interval [s] & $\mathcal U(1,3)$ \\
Linear velocity offset $x,y$ [m/s] & $\mathcal U(-0.5,0.5)$ \\
Linear velocity offset $z$ [m/s] & $\mathcal U(-0.2,0.2)$ \\
Angular velocity offset $x,y$ [rad/s] & $\mathcal U(-0.52,0.52)$ \\
Angular velocity offset $z$ [rad/s] & $\mathcal U(-0.78,0.78)$ \\
\midrule
\rowcolor{black!8}
\multicolumn{2}{l}{Camera reset: student only} \\
\midrule
Focal length scale $(f_x,f_y)$ & $\mathcal U(0.98,1.02)$ \\
Principal point offset [pixel] & $\mathcal U(-2,2)$ \\
Camera position offset, each axis [m] & $\mathcal U(-0.025,0.025)$ \\
Camera rotation offset, each axis [deg] & $\mathcal U(-1,1)$ \\
\midrule
\rowcolor{black!8}
\multicolumn{2}{l}{Observation noise: student only} \\
\midrule
Angular velocity [rad/s] & $\mathcal U(-0.2,0.2)$ \\
Projected gravity & $\mathcal U(-0.05,0.05)$ \\
Joint position [rad] & $\mathcal U(-0.01,0.01)$ \\
Joint velocity [rad/s] & $\mathcal U(-0.5,0.5)$ \\
\bottomrule
\end{tabular*}

\endgroup
\end{table}

\begin{table}[!t]
\centering
\caption{Training hyperparameters and network architecture.}
\label{tab:training-hyperparameters}
\begingroup
\fontsize{8}{9}\selectfont
\setlength{\tabcolsep}{3pt}
\renewcommand{\arraystretch}{1.08}
\begin{tabular*}{\columnwidth}{@{}>{\raggedright\arraybackslash}p{0.40\columnwidth}>{\centering\arraybackslash}p{0.29\columnwidth}>{\centering\arraybackslash}p{\dimexpr0.31\columnwidth-12pt\relax}@{}}
\toprule
Setting & Teacher (PPO) & Student (DAgger) \\
\midrule
\rowcolor{black!8}
\multicolumn{3}{l}{Training hyperparameters} \\
\midrule
Physics / control step [s] & 0.005 / 0.020 & 0.005 / 0.020 \\
Parallel environments & $4 \times 16{,}384$ & $4 \times 16{,}384$ \\
Initial learning rate & $10^{-3}$ & $10^{-3}$ \\
Learning-rate schedule & Adaptive (target KL $0.01$) & Constant \\
Rollout steps per environment & 24 & 16 \\
Optimization epochs per iteration & 5 & 2 \\
Minibatches per epoch & 4 & 8 \\
Gradient norm limit & 1.0 & 1.0 \\
Discount factor $\gamma$ & 0.99 & --- \\
GAE parameter $\lambda$ & 0.95 & --- \\
PPO clipping threshold & 0.2 & --- \\
Value-loss coefficient & 1.0 & --- \\
Clipped value loss & Enabled & --- \\
Entropy coefficient & 0.005 & --- \\
Symmetry loss coefficient & 1.0 & --- \\
Supervised loss & --- & MSE \\
\midrule
\rowcolor{black!8}
\multicolumn{3}{l}{Network architecture} \\
\midrule
Actor MLP & \shortstack{2048, 2048, 1024,\\1024, 512, 512} & \shortstack{1024, 1024, 512,\\512, 256} \\
Critic MLP & \shortstack{2048, 2048, 1024,\\1024, 512, 512} & --- \\
Activation & ELU & ELU \\
Running observation normalization & Disabled & Disabled \\
Depth CNN channels & --- & 32, 32, 64 \\
Depth CNN kernels / strides & --- & $(3,3,3)$ / $(2,2,1)$ \\
Depth CNN pooling & --- & Global max pooling \\
\bottomrule
\end{tabular*}
\endgroup
\end{table}

\textbf{Optimization and architecture.}
Table~\ref{tab:training-hyperparameters} summarizes the training hyperparameters and network architectures of the teacher and student.

\textbf{Actuation.}
The policy action specifies scaled offsets from the nominal joint positions, which a PD controller tracks:
\begin{equation}
\begin{aligned}
\mathbf q_t^{\mathrm{des}} &= \mathbf q^0+\mathbf s\odot\mathbf a_t,\\
\boldsymbol\tau_t &= \mathbf K_p\odot(\mathbf q_t^{\mathrm{des}}-\mathbf q_t)
-\mathbf K_d\odot\dot{\mathbf q}_t.
\end{aligned}
\end{equation}
Here, $\mathbf K_p$ and $\mathbf K_d$ are joint PD gains, and applied torques are limited componentwise to $\pm\boldsymbol\tau^{\max}$. We follow BeyondMimic~\cite{liao2025beyondmimic} for the PD gains, simulated joint armature, torque limits, and action scales $\mathbf s=0.25\boldsymbol\tau^{\max}\oslash\mathbf K_p$, where $\odot$ and $\oslash$ denote elementwise multiplication and division.

\textbf{Adaptive sampling.}
Motion sequences are divided into 50-frame bins. Cumulative failure and visit counts are initialized to one per bin. Failure rates are capped at 200 times their mean, normalized, and mixed with a uniform distribution of weight $0.1$. Bin-length and sequence-length corrections are then applied before renormalization. After selecting a frame within a sampled bin, we subtract a uniformly sampled offset of 0--199 frames, clamped at the sequence start, to practice transitions into difficult segments.

\subsection{Evaluation Details}
\label{sec:appendix-evaluation}

\textbf{Controller comparison.}
The main comparison evaluates 160 motions on flat ground and eight non-flat terrain conditions: ascending and descending stairs and slopes at two levels. Within each condition, metrics are averaged over motions in each of the three motion groups, then equally over groups. Terrain results are averaged equally over the eight conditions. Pose errors use each controller's pre-termination window. Contact F1 compares foot contacts against the same flat-ground reference labels for all controllers, including terrain trials.

\textbf{Ablation protocol.}
For the ablation study in Sec.~\ref{sec:exp-ablation}, the held-out set contains 415 locomotion clips totaling 3,601.72 seconds. Each policy is evaluated on all clips across ascending and descending stairs and slopes at five levels, giving 8,300 trials and 20.01 hours of reference motion per policy.
Step heights are 5, 7.5, 10, 12.5, and 15\,cm, with corresponding slope gradients of 0.15, 0.225, 0.30, 0.375, and 0.45; stair treads are 30\,cm deep. The middle level matches the training terrain scale.

For each metric, we first average equally over clips and the four terrain types within each level. Table~\ref{tab:component-ablations} reports the mean and sample standard deviation of these five level means, rather than variability across training seeds. All trials are retained without outcome-based filtering. Pose errors use each policy's pre-termination window; Contact F1 uses the adapted reference labels. The common-window check truncates each matched trial at the earliest termination among the three policies.

\end{document}